\documentclass[conference,compsoc]{IEEEtran}

\usepackage{multirow}
\usepackage{booktabs}
\usepackage{subfigure}
\usepackage{algorithm}
\usepackage{array}
\usepackage{pifont}
\usepackage{graphicx}
\usepackage{float}
\usepackage{bbm}
\usepackage{spverbatim}
\usepackage{bm}
\usepackage{nicefrac}       
\usepackage{amsmath} 
\usepackage{enumitem}
\usepackage{amssymb}        
\usepackage{caption}        
\usepackage{mathtools}      
\usepackage{amsfonts}
\usepackage{mathtools}
\usepackage{pifont}
\usepackage{xcolor}
\usepackage{xurl}
\usepackage{balance}
\usepackage[flushleft]{threeparttable}
\definecolor{mycitecolor}{RGB}{0,0,153}
\definecolor{mylinkcolor}{RGB}{179,0,0}
\definecolor{myurlcolor}{RGB}{255,0,0}
\definecolor{darkgreen}{RGB}{0,170,0}
\definecolor{darkred}{RGB}{200,0,0}

\usepackage[colorlinks,
            linkcolor=mylinkcolor,
            urlcolor=teal,
            anchorcolor=blue,
            citecolor=mycitecolor,
            ]{hyperref}

\setlist[itemize]{leftmargin=*,noitemsep, topsep=0pt}

\usepackage{xspace}
\usepackage{multirow}
\usepackage{amsthm}
\theoremstyle{plain}
\newtheorem{definition}{Definition}

\usepackage{dsfont}

\usepackage[noend]{algpseudocode}

\makeatletter
\newenvironment{breakablealgorithm}
  {
    \par\vspace{\intextsep}
      \refstepcounter{algorithm}
      \hrule height.8pt depth0pt \kern2pt
      \renewcommand{\caption}[2][\relax]{
        {\raggedright\textbf{\fname@algorithm~\thealgorithm} ##2\par}%
        \ifx\relax##1\relax 
          \addcontentsline{loa}{algorithm}{\protect\numberline{\thealgorithm}##2}%
        \else 
          \addcontentsline{loa}{algorithm}{\protect\numberline{\thealgorithm}##1}%
        \fi
        \kern2pt\hrule\kern2pt
     }
  }{
      \kern2pt\hrule\relax
    \par\vspace{\intextsep}
  }
\makeatother

\newcommand{\ie}{\emph{i.e., }}
\newcommand{\eg}{\emph{e.g., }}

\newcommand{\aka}{\emph{aka. }}

\newcommand{\mypara}[1]{\smallskip\noindent\textbf{#1.} \xspace}

\newcommand\mc{\mathcal}
\newcommand{\eps}{\epsilon}
\newcommand{\norm}[1]{\left\lVert#1\right\rVert}
\newcommand{\pb}[1]{%
    \parbox[t]{\dimexpr\linewidth-\algorithmicindent}{#1\strut}%
}
\renewcommand{\colon}{\nobreak\mskip2mu\mathpunct{}\nonscript
  \mkern-\thinmuskip{:}\mskip6muplus1mu\relax}

\DeclareCaptionFormat{algor}{%
  \hrulefill\par\offinterlineskip\vskip1pt%
    \textbf{#1#2}#3\offinterlineskip\hrulefill}
\DeclareCaptionStyle{algori}{singlelinecheck=off,format=algor,labelsep=space}
\title{Private Linear Regression with Differential Privacy and PAC Privacy}

\author{
\IEEEauthorblockN{Hillary Yang}
\IEEEauthorblockA{
Carmel High School \\
hillaryyang2@gmail.com}
\and
\IEEEauthorblockN{Yuntao Du}
\IEEEauthorblockA{
Department of Computer Science\\
Purdue University\\
ytdu@purdue.edu}
}

\begin{document}
\maketitle

\begin{abstract}
Linear regression is a fundamental tool for statistical analysis, which has motivated the development of linear regression methods that satisfy provable privacy guarantees so that the learned model reveals little about any one data point used to construct it.
Most existing privacy-preserving linear regression methods rely on the well-established framework of differential privacy, while the newly proposed PAC Privacy has not yet been explored in this context.
In this paper, we systematically compare linear regression models trained with differential privacy and PAC privacy across three real-world datasets, observing several key findings that impact the performance of privacy-preserving linear regression.
\end{abstract}

\section{Introduction}

Differential privacy (DP) is a widely accepted privacy notion. Rooted in the input-independent indistinguishability established in Shannon's perfect secrecy and cryptography, DP measures the posterior advantage, \ie the difference between the prior and posterior success rate for an adversary to identify the membership in a dataset in the worst case. The worst case or input independence here refers to such posterior advantage characterization holding for arbitrary prior data distribution. Despite being a powerful privacy guarantee, DP also encounters several fundamental challenges in application. From an operational perspective, one major issue is that most DP randomization approaches, for example, Gaussian/Laplace mechanisms and exponential mechanism \cite{dwork2014algorithmic}, require determining the sensitivity, \ie the worst-case change that arbitrarily replaces a single data point can produce, of an algorithm,  but in general tight sensitivity is known to be NP-hard to compute \cite{sensitivity-NPhard}. In many practical applications, sensitivity can also be loosely upper-bounded and it is even challenging to conduct DP analysis on black-box processing functions.

To address the challenges in black-box analysis, another privacy framework, called PAC Privacy \cite{xiao2023pac}, is proposed. PAC Privacy models the information leakage as an inference task to challenge the adversary to return a satisfied estimation on any predefined sensitive feature of the input, where the security parameter is selected as the adversary's posterior success rate. Instead of adopting an input-independent setup, PAC Privacy assumes and exploits the input entropy, i.e., considering the input is generated from some (possibly black-box) distribution. PAC Privacy establishes a set of new information theoretical tools, as an end-to-end perturbation, to efficiently and automatically determine the noise to provably control the adversarial posterior success rate through black-box evaluations of objective processing function.

In this work, we study the problem of privacy-preserving linear regression using differential privacy and PAC privacy. 
Extensive research has been conducted on differentially private linear regression, including approaches such as objective perturbation~\cite{kifer2012private}, ordinary least squares (OLS) with noisy sufficient statistics~\cite{dwork2014analyze,wang2018revisiting}, and DP-SGD~\cite{abadi2016deep}. 
Recent studies~\cite{amin2022easy} have shown that when properly tuned, the linear regression trained with DP-SGD (\aka DPSGD-LR) can outperform many sophisticated differentially private algorithms for linear regression.
On the other hand, adopting PAC privacy for linear regression has been less explored. 
There is also a lack of head-to-head comparison between differentially private and PAC private linear regression methods, and a comprehensive understanding of these privacy notions’ strengths and weaknesses on linear regression remains elusive.
To address this gap, this work conducts a series of experiments on privacy-preserving linear regression under these two privacy notions and finds several interesting findings.
Specifically, our main contributions are as follows:
\begin{itemize}
    \item We introduce a new private linear regression algorithm with PAC privacy, called PAC-LR, which estimates anisotropic noise for membership privacy and can be directly compared with DP under the same criteria.
    \item We conduct a series of experiments on three real-world datasets to compare PAC-LR and DPSGD-LR, finding that PAC-LR outperforms DPSGD-LR, particularly under strict privacy guarantees.
    \item We observe that data normalization significantly impacts both DPSGD-LR and PAC-LR, and regularization techniques are essential for making PAC-LR more robust and effective.
\end{itemize}

\section{Related work}


\subsection{Differentially Private Linear Regression}

Extensive studies have been explored on differentially private linear regression. 
Based on convex optimization, Chaudhuri et. al~\cite{chaudhuri2011differentially} propose objective perturbation, which perturbs the objective function that is the target of optimization. Performing regression using this algorithm is quite restrictive, as it requires an exact minimum can be found, which is not feasible in practice, especially considering iterative optimizers; notably, gradient descent. However, recent work has loosened these restrictions to finding approximate minima~\cite{iyengar2019towards}, and lessened noise~\cite{kifer2012private}, and have produced highly accurate empirical results.

Due to varying opinions on the effectiveness and utility of output perturbation \cite{yu2019gradient, wang2017differentially}, attention has turned to differentially private stochastic gradient descent (DPSGD) as the favored approach for learning tasks (\cite{song2013stochastic, bassily2014private, abadi2016deep, ICML2020,hu2022high,xiao2019local,xiao2022differentially}). By perturbing each calculated gradient accordingly to limit sensitivity, differential privacy can be achieved by the compositional property \cite{dwork2014algorithmic}. However, DPSGD does experience drawbacks, including the computation and memory overheads on per-sample gradient clipping \cite{CCS2023}, privacy costs of training (hyperparameter tuning), and the significant accuracy losses and biases \cite{SP2023} incurred after making a mechanism differentially private \cite{bagdasaryan2019differential}.

Other techniques for differentially private linear regression include sufficient statistics perturbation \cite{vu2009differential, wang2018revisiting, alabi2020differentially}, which achieves DP by releasing noisy sufficient statistics and can display optimal performance for varying parameter settings \cite{wang2018revisiting}. Amin et. al \cite{amin2022easy} develop TukeyEM for linear regression using a propose-test-release (PTR) framework \cite{dwork2009differential}, which does not require data bounds and hyperparameter tuning. Only one parameter is chosen by the user, namely, the number of models, so no privacy is lost to instance-specific inputs such as feature and label norms. However, for the PTR check to pass, and a model to be output, this algorithm requires $n \gtrsim 1000d$ for a dataset of dimension $d$ and size $n$. Therefore, TukeyEM is impractical for a wider range of datasets, and has been shown to achieve suboptimal results \cite{dick2024better}.

\subsection{PAC Privacy}

Newly proposed Probably Approximately Correct (PAC) Privacy \cite{xiao2023pac}, effectively eliminates the computation and proof of DP's worst-case sensitivity with an automatic privatization algorithm, detailed by Sridhar et. al \cite{sridhar2024pac}. Further contrasting to the disproportionately large amounts of noise required of DP, the noise magnitudes of PAC can be $\mc{O}(1)$ in some cases.

Sridhar et. al \cite{sridhar2024pac} perform qualitative, empirical analyses of PAC Privacy for machine learning techniques, suggesting both a template for automatic privatization and an efficient algorithm for anisotropic noise generation that emphasizes the benefits of algorithm stability. Using four common machine learning algorithms, ranging from K-Means clustering to classification with SVM, it was found that privacy-utility tradeoffs can be greatly improved, especially for algorithms that only experience ``superficial" instabilities --- differences in the output that do not alter functionality or interpretation --- such as K-Means. 

Another consideration is how easily an algorithm can be modified to fit the PAC framework. K-Means can be easily adapted, with the observed output being the centroids, to which we want to minimize the sum-of-squares from the sensitive input, the values in each cluster \cite{arthur2006k}. However, random forest algorithms face more difficulty. In particular, PAC requires an ordered list of the features that each tree can split into, which defines the same structure for every tree. Furthermore, to ensure stability, regularization and data augmentation techniques must be employed. Similar requirements for modification are faced by DP machine learning algorithms \cite{su2016differentially}.


\section{Preliminaries}

\subsection{Differential Privacy}
\begin{definition}[$(\eps, \delta)$-differential privacy \cite{dwork2006differential}]
    Define a randomized mechanism $\mc{M}\colon \mc{X} \to \mc{Y}$. Then, for two datasets $D, D' \subseteq \mc{X}$ that differ by the presence or absence of a single record, $\mc{M}$ is $(\eps, \delta)$-differentially private if for any $y \subseteq \mc{Y}$
    \begin{equation}
        \Pr(\mc{M}(D) \in y) \leq e^{\eps} \cdot \Pr(\mc{M}(D') \in y) + \delta
    \end{equation}
    where $\eps \ge 0$ and $\delta \in [0, 1]$.
\end{definition}

Intuitively, differential privacy aims to reduce the effect of any individual's data on the output of a data processing function. The parameter $\eps$, the \textit{privacy budget}, serves as a quantification of the probability of information disclosure, and thus, the potential harm of the outputs. Smaller $\eps$ indicates a stronger privacy guarantee but detracts from utility \cite{dwork2014algorithmic}. The second parameter, $\delta$, describes the probability that a leakage occurs, which happens when the added noise is not sufficient for privacy. Typically, we choose $\delta < 1/n$, for a dataset with $n$ records. 



\subsection{PAC Privacy}
\begin{definition}[$(\delta, \rho, D)$-PAC Privacy \cite{xiao2023pac,CCS2024}]
    With a processing mechanism $\mc{M}$, from a data distribution $D$, and a measure function $\rho(\cdot, \cdot)$, $\mc{M}$ is defined to satisfy $(\delta, \rho, D)$-PAC Privacy if the following experiment is impossible:

    A user generates data $X$ from a distribution $D$. Then, $\mc{M}(X)$ is revealed to an informed adversary, who returns an estimation $\hat{X}$ on $X$. The probability that $\rho(\hat{X}, X) = 1$ is at least $1-\delta$.
\end{definition}


PAC provides a solution to the unaddressed questions of DP; namely, its unintuitive, probabilistic guarantees and lack of concrete privacy metrics \cite{lee2011much}. On the other hand, PAC addresses and formalizes reconstruction hardness by providing a concrete upper bound on the adversary's posterior success rate, $1-\delta$. 


To align with the existing privacy notions like differential privacy, we focus on the special case of PAC privacy: membership privacy~\cite{ccs13membership}.
Specifically, let $\bm{1}_i$ represent an indicator which equals $1$ if the $i$-th datapoint $u_i$ from the entire data pool $\mathsf{U}=\{u_1, u_2, \cdots, u_n\}$ is included in the sampled set $X$ otherwise $0$. The adversary aims to recover $\bm{1}_i$ by some estimate $\hat{\bm{1}}_i$. As a corollary from \cite{CCS2024}, under a Poisson sampling $q$, the mutual information $\mathsf{MI}\big(\bm{1}_i,\mathcal{M}(X)\big)$ can be upper bounded as
\begin{multline*}
\mathsf{MI}\big(\bm{1}_i,\mathcal{M}(X)\big) \leq q(1-q)\mathbb{E}_{X_{-i}} \big[\mathcal{D}_{KL}(\mathbb{P}_{\mathcal{M}(X_{-i},u_i)}\| \mathbb{P}_{\mathcal{M}(X_{-i})}) \\
+\mathcal{D}_{KL}(\mathbb{P}_{\mathcal{M}(X_{-i},u_i)}\| \mathbb{P}_{\mathcal{M}(X_{-i})}\big]
\label{individual_MI}
\end{multline*}

Here, $X_{-i}$ represents a random set produced by conducting a Poisson sampling of rate $q$ over $\mathsf{U}\slash u_i$. In particular, when we consider adding (multivariate) Gaussian noise to some deterministic processing function $\mathcal{F}(\cdot)$, i.e., $\mathcal{M}(X)=\mathcal{F}(X)+e$ for $e \sim \mathcal{N}(0,\Sigma)$, then 
\begin{equation*}
\begin{aligned}
&\mathsf{MI}\big(\bm{1}_i,\mathcal{M}(X)\big) \leq q(1-q)\mathbb{E}_{X_{-i}} \big[ \|\mathcal{F}(X_{-i},u_i)-\mathcal{F}(X_{-i})\|^2_{\Sigma^{-1}}\big]
\end{aligned}
\label{individual_MI1}
\end{equation*}

\section{Private Linear Regression}

\subsection{Linear Regression}


Linear regression is a statistical method for modeling the relationship between a dependent or response variable $y$ and one or more independent variables $x_i$. The relationship has the form:
\begin{equation*}
  y = \mathbf{X}\beta + \eps
\end{equation*}

where $\mathbf{X}$ is the feature matrix with $n$ features, $\beta$ is the vector of regression coefficients, and $\eps$ is the vector of the model's error terms.

The ordinary least squares (OLS) method is commonly used to fit the model, by choosing coefficients such that 

\begin{equation*}
    S(\beta) = \sum_{i = 1}^n \Big(y_i - \beta^\top x_i\Big)^2 = \norm{\mathbf{y} -  \mathbf{X} \beta}_2^2
\end{equation*}

is minimized. The least-squares solution is also given by solving the normal equations: 

\begin{equation*}
(\mathbf{X}^\top \mathbf{X}) \beta = \mathbf{X}^\top \mathbf{y}
\end{equation*}

where $\mathbf{y}$ is the vector of predicted values.

\mypara{Regularization}
When multicollinearity is present in a dataset, linear regression may become unstable and lose accuracy. Ridge and Lasso regression correct for these errors by including penalty terms to the loss function. These are the $l_2$ and $l_1$ norms of the coefficients, for Ridge and Lasso, respectively.

Ridge, or $l_2$, regression shrinks the coefficients towards zero, adding a penalty proportional to the sum of the squares of the coefficients. This regularization term can be seen in the minimization of:

\begin{equation*}
    \norm{\mathbf{y} - \mathbf{X}\beta}_2^2 + \lambda \norm{\beta}_2^2
\end{equation*}

where $\lambda \ge 0$ is the regularization parameter controlling the penalty strength. Explicitly, the solution $\beta$ that minimizes this is:

\begin{equation*}
\beta = (\mathbf{X}^\top \mathbf{X} + \lambda \mathbf{I})^{-1}\mathbf{X}^\top\mathbf{y}
\end{equation*}

Lasso regression is similar in principle, but tends to lead to sparsity, pushing some of the coefficients to zero when $\lambda$ is large. This is useful for feature selection, where the most important features are identified. The objective is to minimize:

\begin{equation*}
\norm{\mathbf{y} - \mathbf{X}\beta}_2^2 + \lambda\norm{\beta}_1
\end{equation*}

\subsection{Linear Regression with DPSGD}

After machine learning models have become pervasive, so has a need for securing the data that goes into training them. Differentially private stochastic gradient descent (DPSGD-LR) is the standard method for privatizing linear regression, as shown below.

\begin{breakablealgorithm}
\caption{Differentially Private Stochastic Gradient Descent} \label{dpsgd}
\textbf{Input:} Training features $\{x_1, \dots, x_n\}$, loss function $\mc{L}(w_i, x_i)$ for parameters $w_i$, epsilon $\eps$ and delta $\delta$\\
\textbf{Hyperparameters}: Learning rate $\eta$, batch size $b$, epochs $T$, norm clip $C$, noise multiplier $\sigma$
\begin{algorithmic}[1]
\State \textbf{Initialize} weights $w_0$ randomly
\For{$t \in [T]$} 
    \State \pb{Randomly sample batch $B$ of size $b$} 
    \State \pb{\textbf{Compute gradients} $g_t(x_i)$ for $x_i \in B$} 
    \State \pb{\textbf{Clip gradient} \\ $\hat{g}_t(x_i) \leftarrow g_t(x_i) / \max(1, \norm{g_t(x_i)}_2/C)$} 
    \State \pb{\textbf{Aggregate and perturb} \\ $\bar{g}_t \leftarrow \frac{1}{b}\sum_{i} \hat{g}_t(x_i) + \mc{N}(0, \sigma^2C^2 \mathbf{I})$} 
    \State \pb{\textbf{Update:} $w_{t+1} \leftarrow w_t - \eta \bar{g}_t$} 
\EndFor
\end{algorithmic}
\textbf{Output:} Model weights $w_T$
\end{breakablealgorithm}

Similar to regular SGD, at epoch $t$, a mini-batch of size $b$ is sampled uniformly at random from the dataset. (line 3 in Algorithm \ref{dpsgd}) The gradients $\hat{g}(x)_i \leftarrow \nabla_{w_t} \mc{L}(w_t, x_i)$ are then computed for each data point using the specified loss function $\mathcal{L}$ (line 4). 

DPSGD-LR incorporates two key mechanisms on top of the traditional SGD algorithm:
\begin{enumerate}
    \item \textbf{Gradient clipping.} (line 5) Each per-example gradient is scaled to ensure that its $\ell_2$-norm does not exceed a predefined threshold $C$, bounding the sensitivity of gradient updates and limiting the maximum impact of individual data points.
    \item \textbf{Noise addition.} (line 6) At each iteration, Gaussian noise proportional to $C$ is added to the gradients. The noise level is determined by the privacy parameters $\eps$ and $\delta$. More noise yields a stronger privacy guarantee at the expense of model utility. 
\end{enumerate}

These modifications ensure that the resulting trained model adheres to the guidelines of DP, with privacy loss accumulating over multiple iterations via the composition property of DP. Along with a specification of $\eps$, this forms the basis of a powerful technique used for training machine learning models.

\subsection{Linear Regression with PAC Privacy}

\mypara{Calculating Variance and Noise for PAC}
PAC requires the user to specify the desired maximal posterior success rate, given as a decimal on $(0.50, 1.00)$. We can translate this posterior advantage into a quantification of mutual information using Equation (3) in Sridhar et. al \cite{sridhar2024pac}, given by:
\[
p_o \ln\Big(\frac{p_o}{p}\Big) + (1 - p_o) \ln\Big(\frac{1-p_o}{1-p}\Big) \leq \text{MI}
\]
where $p$ is the prior success rate and $p_o$ the posterior. For the membership inference task, $p_o = 0.50$, so 
\[
\text{MI} \geq p_o \ln(2p_o) + (1 - p_o) \ln(2-2p_o)
\]
Thus, we have an expression for the lower bound on the mutual information, given the maximal posterior success rate. 

\mypara{Basis optimization with SVD} 
The noise levels can be further optimized by calculating the singular-value decomposition (SVD) of the outputted weights, as shown in Algorithm 1 of \cite{sridhar2024pac}. We can then decompose the weight matrix $A$ as the product $U\Sigma V^T$, for unitary matrices $U$ and $V^T$. By multiplying $y = V^Tx$, $x$ is projected onto a new basis defined by the columns of $V$. In other words, the coordinate axes are transformed so that they align with the directions of maximum variance. The noise calculated according to this new basis leads to smaller sensitivity and thus, optimized noise. 

Using the MI value and $V^T$ as a unitary projection matrix, we can estimate the anisotropic noise levels required for membership privacy with the following algorithm. 

\begin{breakablealgorithm}
\label{alg:pac_noise_learn}
\caption{Anisotropic noise estimation for PAC membership privacy}
\textbf{Input}: Data $D$ with size $n$, dimension $d$, black-box mechanism $\mc{M}$, mutual information MI, convergence threshold $\eta$, $d \times d$ projection matrix $V^T$
\begin{algorithmic} 
\State \textbf{Initialize} \textit{converged} = \textbf{False}, $v$ as a map to $d$ lists, $e_{max}$ to store max noise in each dimension
    \For{$x \in D$}
        \State $D' = D \setminus \{x\}$
        \While{\textbf{not} \textit{converged}}
            \State \pb{
            $A: $ Poisson sampling half from $D'$ \\
            $B = D' \cup x$} \\

            \State \pb{
            $M^\prime(A) \leftarrow V^T \cdot M(A)$ \\
            $M^\prime(B) \leftarrow V^T \cdot M(B)$
            } \\
            
            \State \pb{
            $g = \norm{M^\prime(A) - M^\prime(B)}_2$ \\
            Append $g[i]$ to $v[i]$ for $i \in [d]$ \\
            } 
            \If {mean of $v[i] - $ previous mean $<\eta$} 
                \State \textit{converged} = \textbf{True}
            \EndIf
        \EndWhile
        \State \pb{\textbf{Compute noise}
        \begin{equation*}
        \text{For $i \in [d]$, } e[i] \leftarrow \frac{\sqrt{v[i]} \sum_{j=1}^d \sqrt{v[j]}}{4\cdot \text{MI}}
        \end{equation*}

        \textbf{Project to original space} \\
        $d \times d$ diagonal matrix $\Sigma^\prime$: \\
        $\Sigma^\prime[i][i] \leftarrow e[i]$ for $i \in [d]$ \\
        $\Sigma = V^T \cdot \Sigma^\prime$ \\
        $e[i] \leftarrow \Sigma[i][i]$ for $i \in [d]$ \\
        
        $e_{max}[i] = \max(e[i], e_{max}[i])$ for $i \in [d]$
        }
    \EndFor
\end{algorithmic}
\textbf{Output:} Maximum noise for each output dimension, $e_{max}$
\end{breakablealgorithm}

Once the noise has been estimated, we privatize the outputs of $\mc{M}$. In the case of linear regression, this output is the model's weights, including its intercept. After transforming the weight vector back to the original space, we add Gaussian noise drawn randomly from $\mc{N}(0, e[i])$ to each output dimension as shown below.

\begin{breakablealgorithm}
\caption{Anisotropic noise addition}
\label{alg:pac_noise_add}
\textbf{Input}: Trained model parameters $w$ with dimension $d$ and learned noise $e$ from Algorithm \ref{alg:pac_noise_learn}
\begin{algorithmic}
\For{$i \in [d]$} 
    \State \pb{
    $c \sim \mc{N}(0, e[i])$
    }
    \State \pb{Add noise to parameters \\
    $w[i] += c$
    }
\EndFor
\end{algorithmic}
\label{alg3}

\textbf{Output:} Noisy model parameters
\end{breakablealgorithm}

\subsection{Comparing DP and PAC}
For some mechanism satisfying $(\eps, \delta)$-DP, Humphries et. al \cite{humphries2023investigating} show that the posterior success rate $p_o$ for membership inference is upper bounded by 
\begin{equation*}
p_o \leq 1 - \frac{1-\delta}{1 + e^\eps}
\end{equation*}
which gives us a lower bound on epsilon:
\begin{equation*}
\eps \geq \ln\Big(\frac{1-\delta}{1-p_o} - 1\Big).
\end{equation*}
The calculated relationships between various posterior success rates (PSR), mutual information, and epsilon are shown below. These are also the values we used in our experiments.
\begin{table}[t]
    \centering
    \caption{Connecting DP (privacy budget $\epsilon$) and PAC privacy (mutual information, MI) with posterior success rate.}
    {\small \resizebox{0.46\textwidth}{!}{
\begin{tabular}{lccccccccc}
\toprule
\textbf{PSR} & \textbf{Privacy Budget } $\epsilon$ & \textbf{Mutual Information (MI)}   \\
\midrule
0.52 & 0.080023 & 0.000800  \\
\midrule
0.55 & 0.200652  & 0.005008 \\
\midrule
0.65 & 0.619023 & 0.045700  \\
\midrule
0.75 & 1.098598 & 0.130812  \\
\midrule
0.85 & 1.734589 & 0.270438  \\
\midrule
0.95 & 2.944428 & 0.494631 \\
\midrule
0.98 & 3.891810 & 0.595108  \\
\bottomrule
\end{tabular}}}
    \label{tab:connect_dp_pac}
\end{table}

\section{Experiments}

\subsection{Experimental Setups}

\mypara{Datasets}
We utilized three real-world datasets for our experiments, each differing in size, characteristics, attributes, and distribution.
These datasets were intentionally selected for their small size to facilitate performance comparisons across different privacy notions, as it is generally believed that small datasets tend to yield poor utility when subjected to provable privacy guarantees.
Additionally, the inherent characteristics of the datasets make them well-suited for linear regression.
Table~\ref{tab:dataset_info} presents a statistical overview of these datasets, with detailed descriptions provided below.

\begin{itemize}
    \item \textbf{Lenses}~\cite{lenses_58} contains four features of patients (\ie age, spectacle prescription, tear production rate, and astigmatism)  to predict the type of contact lenses a patient should be fitted with.
    \item \textbf{Concrete}~\cite{concrete_slump_test_182} consists of 103 data samples and 7 features that are used to predict the compressive strength of the concrete. 
    \item \textbf{Automobiles}~\cite{automobile_10} predicts the automobile's price using the car's characteristics, such as the number of doors and miles per gallon. We replaced the missing values with the mean values in that column. 
\end{itemize}

All datasets are publicly available and can be downloaded from the UCI Machine Learning Repository\footnote{\url{https://archive.ics.uci.edu/}}.

\mypara{Evaluation Metrics}
We use the root mean squared error (RMSE) as our primary evaluation metric, as it provides a more direct indicator of model performance.
RMSE can be interpreted as the standard deviation of the prediction errors, where lower RMSE values indicate better performance.
In addition, we utilize the widely used coefficient of determination,  $R^2$, to measure the variation in labels accounted for by the features. 
$R^2=1$ is perfect, $R^2=0$ is the trivial baseline achieved by simply predicting the average label, and $R^2<0$ is worse than the trivial baseline.

\begin{table}[t]
    \centering
    \caption{Summary of used datasets.}
    {\small \resizebox{0.43\textwidth}{!}{
\begin{tabular}{lccccccccc}
\toprule
 & \textbf{\# Size} & \textbf{\# Features}  & \textbf{Task} \\
\midrule
\textbf{Lenses} & $24$ & $4$ & Regression  \\
\textbf{Concrete} & $103$ & $7$ & Regression   \\
\textbf{Automobiles} & $201$ & $15$ & Regression \\
\bottomrule
\end{tabular}}

}
    \label{tab:dataset_info}
\end{table}

\subsection{Linear Regression Algorithms}
We compare the performance of two different types of private linear regression algorithms (\ie differential privacy and PAC privacy), along with the non-private version as follows:
\begin{itemize}
    \item \textbf{Linear Regression with DPSGD (DPSGD-LR).} We implement differentially private linear regression using the Opacus library \cite{opacus}. Opacus provides the \verb|PrivacyEngine| class, which facilitates the privatization of PyTorch training objects. Specifically, we utilize the \verb|make_private_with_epsilon| function, which allows us to specify the desired privacy budget for training the private linear regression model.
    \item \textbf{Linear Regression with PAC Privacy (PAC-LR).} For PAC privacy, we first follow Algorithm \ref{alg:pac_noise_learn} to sample the training data for anisotropic noise esitmations. Once the noise has been learned, we privatize the weights of linear regression by adding the Gaussian noise.
    \item \textbf{Non-private Linear Regression (Non-Private).} To compare the performance with the non-private baseline, we utilize Ordinary Least Squares (OLS) regression from \verb|sklearn| library. OLS determines the parameters of the linear regression model by minimizing the sum of the squared differences between the actual and predicted values, providing a closed-form solution for linear regression.
\end{itemize}

\subsection{Implementation Details}

\mypara{Data Normalization}
In our experiments, we observed that data normalization significantly impacts the performance of privacy-preserving linear regression (as detailed in Section~\ref{sec:exp_albation}).
To address this, we preprocess the datasets using the \verb|StandardScaler| from \verb|sklearn| library, ensuring that each feature is standardized to have a mean of 0 and a standard deviation of 1.

\mypara{Regularization}
We also observe that regularization has a significant impact on private linear regression, particularly when the privacy budget is small.
Therefore, we employ two widely used regularization techniques for comparison: Lasso and Ridge regression, which add $l_1$ and $l_2$ norms to the loss functions during optimization.  
We also use the implementations from \verb|sklearn| library for experiments.

\begin{table}[t]
    \centering
    \caption{Hyperparameter search space.}
    {\small 

\resizebox{0.44\textwidth}{!}{
\begin{tabular}{cccccc}
\toprule[0.8pt]
& \textbf{Parameter} & \textbf{Distribution} \\
\midrule

\multirow{4}{*}{\begin{tabular}[c]{@{}c@{}} DPSGD-LR \end{tabular}} 
& Learning Rate & $\{10^{-8}, \dots, 10^{-1}\}$ \\ 
\cmidrule(lr){2-3} 
& Batch Size & $\{2, 2^2, \dots, 2^5\}$ \\
\cmidrule(lr){2-3} 
& Norm Clip & $\{1, 10^2, \dots, 10^7\}$ \\
\cmidrule(lr){2-3} 
& Epochs & $\{1, 5, 10, 20, 30\}$ \\

\midrule
\multirow{1}{*}{\begin{tabular}[c]{@{}c@{}} PAC-LR \end{tabular}} 
& Regularization Penalty $\lambda$ & $\{2^{-10}, 2^{-9}, \dots, 2^{10}\}$\\ 
\bottomrule
\end{tabular}}}
    \label{tab:hyperparams}
\end{table}

\begin{figure*}[t]
    \centering
    \subfigure[Lenses Dataset]
    {
    \includegraphics[width=0.3\textwidth]{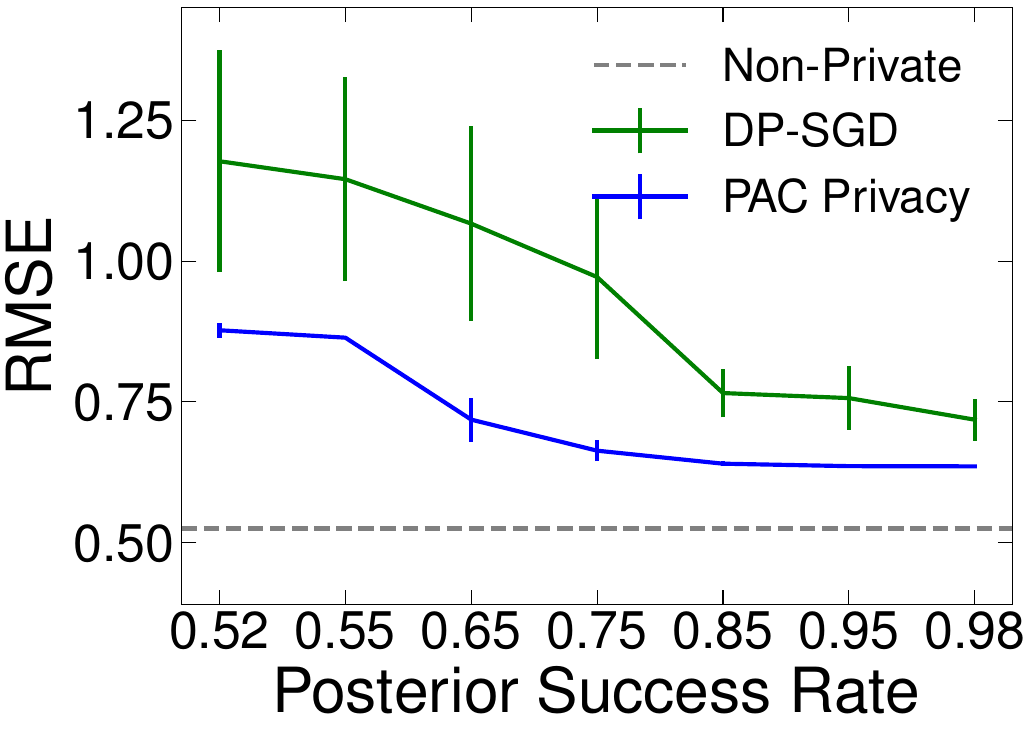}
    \label{fig:lenses}
    }
    \hspace{1.5mm}
    \subfigure[Concrete Dataset]
    {
    \includegraphics[width=0.3\textwidth]{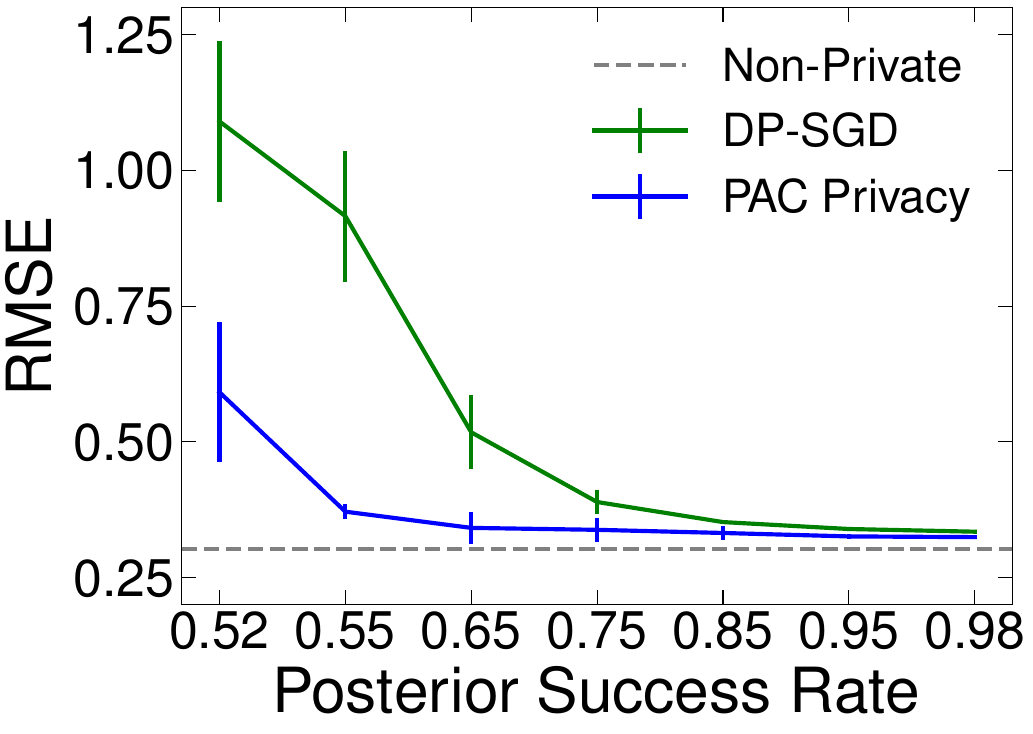}
    \label{fig:concrete}
    }
    \hspace{1.5mm}
    \subfigure[Automobiles Dataset.]
    {
    \includegraphics[width=0.3\textwidth]{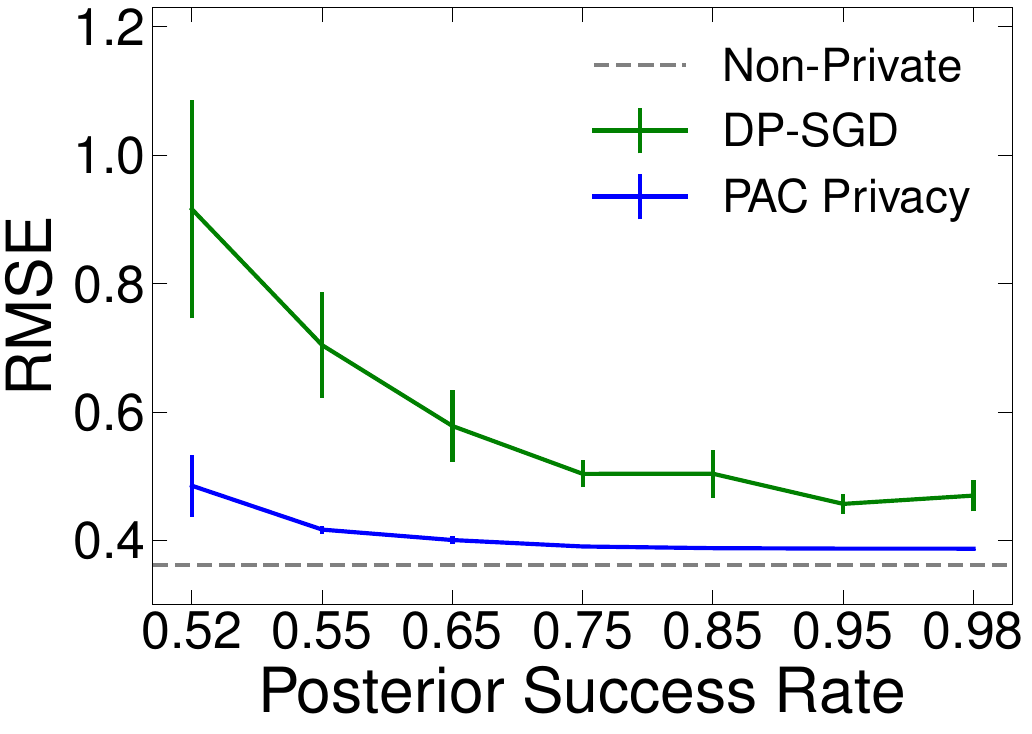}
    \label{fig:automobiles}
    }
    \caption{Performance comparison (\ie RMSE) of private linear regression using DPSGD and PAC privacy across three datasets. The posterior success rate is used to bridge the connection between differential privacy and PAC privacy (see Table~\ref{tab:connect_dp_pac} for details).}
    \label{fig:overall_rmse}
\end{figure*}

\begin{figure*}[t]
    \centering
    \subfigure[Lenses Dataset]
    {
    \includegraphics[width=0.31\textwidth]{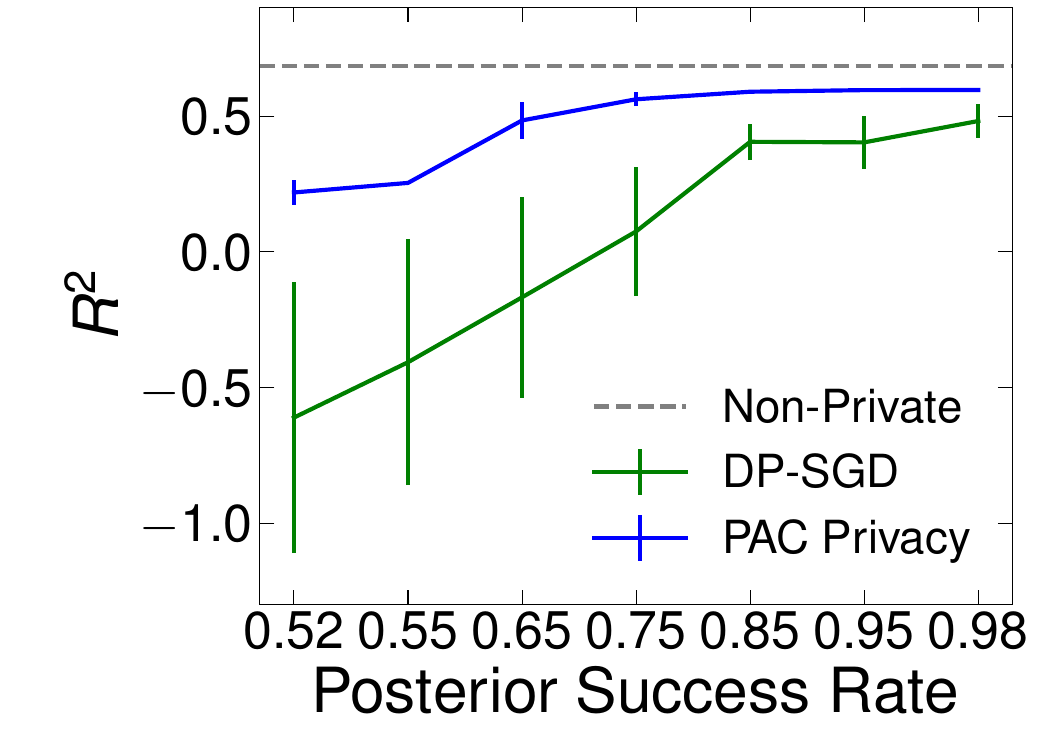}
    \label{fig:r2_lenses}
    }
    \subfigure[Concrete Dataset]
    {
    \includegraphics[width=0.31\textwidth]{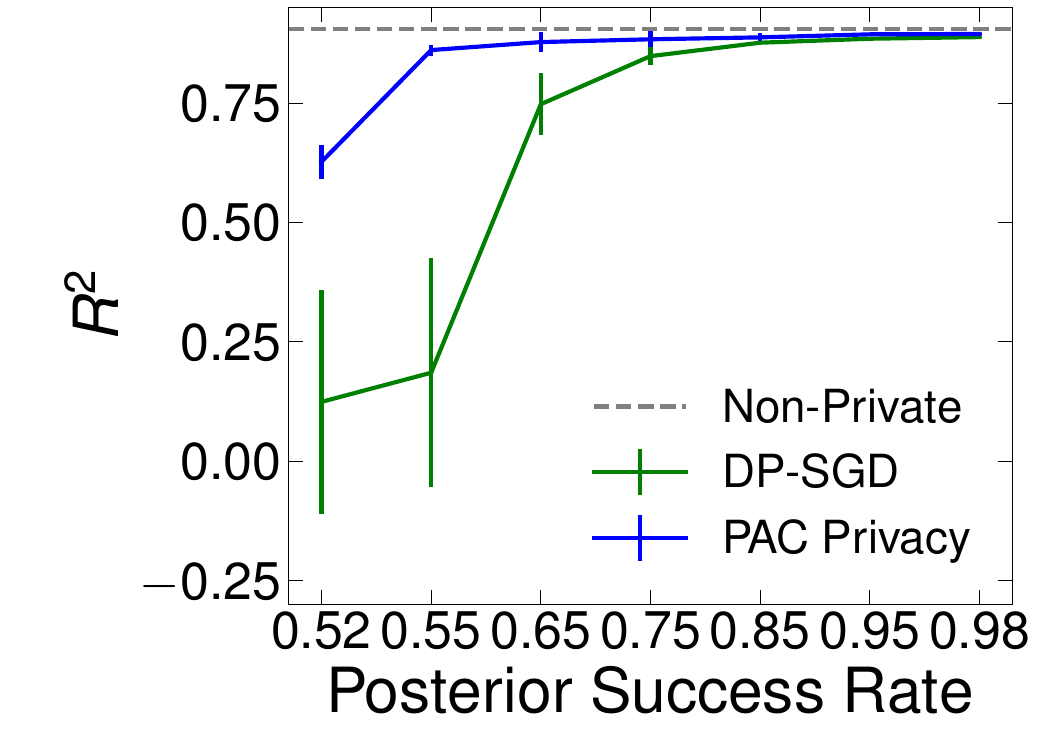}
    \label{fig:r2_concrete}
    }
    \subfigure[Automobiles Dataset.]
    {
    \includegraphics[width=0.31\textwidth]{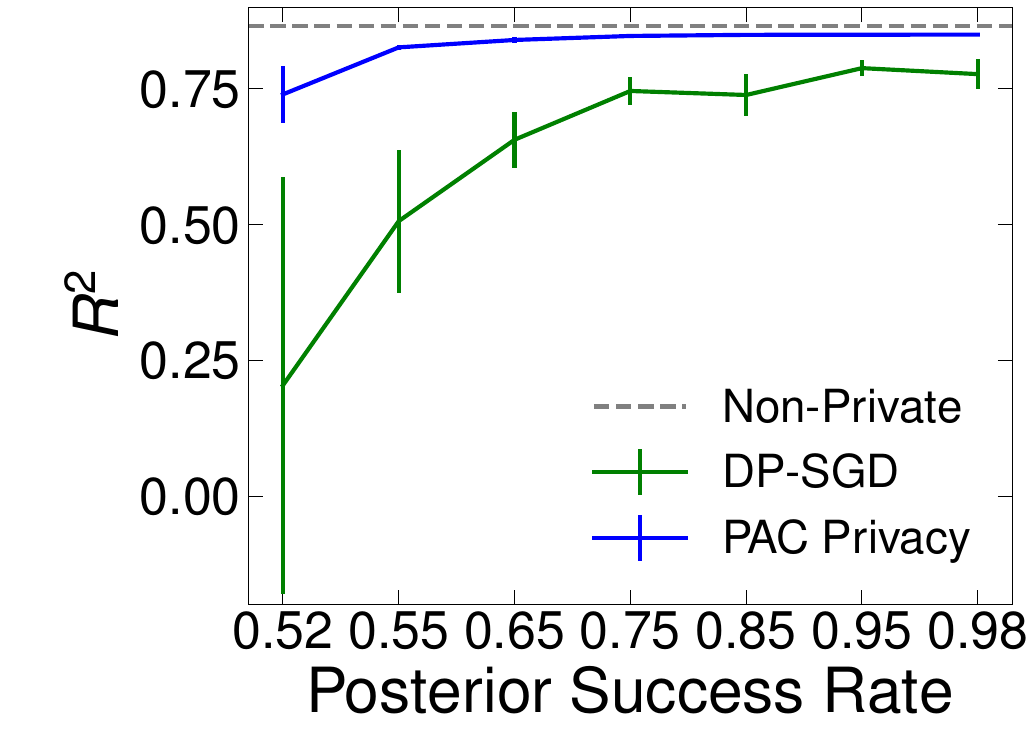}
    \label{fig:r2_automobiles}
    }
    \caption{Performance comparison (\ie $R^2$) of private linear regression using DPSGD-LR and PAC-LR across three datasets. The posterior success rate bridges the connection between differential privacy and PAC privacy (see Table~\ref{tab:connect_dp_pac} for details).}
    \label{fig:overall_r2}
\end{figure*}

\mypara{Hyperparameter Tuning}
We train the private linear regression models 50 times and compute the average and standard deviation of the RMSE and $R^2$ scores for evaluation. 
For PAC Ridge and Lasso, we perform a grid search over the regularization penalty $\alpha$, where  $\alpha \in \{2^{-10}, 2^{-9}, \dots, 2^{10}\}$, and record the value that yields the lowest RMSE.
We also conduct a grid search for the hyperparameters of DPSGD, as outlined in Table~\ref{tab:hyperparams}.

\subsection{Overall Performance}
\label{sec:exp_overall}

We begin with the performance comparison in terms of RMSE and $R^2$.
The experimental results are shown in Figure~\ref{fig:overall_rmse} and Figure~\ref{fig:overall_r2}, with non-private linear regression plotted as dotted lines for reference.
We have the following observations:
\begin{itemize}
    \item  The performance of DPSGD-LR and PAC-LR improves as the posterior success rate increases, with a corresponding decrease in variance. Notably, when the posterior success rate reaches 0.85 for the Concrete dataset, the performances of DPSGD-LR and PAC-LR are very close to that of the non-private model.
    \item PAC-LR demonstrates impressive performance under strict privacy guarantees (\ie small posterior success rates). For instance, when the posterior success rate is low (\eg 0.52), PAC-LR significantly outperforms DPSGD-LR, showcasing the advantage of learning anisotropic noise for private linear regression. In addition, 
    \item While DPSGD-LR shows inferior results compared to PAC-LR, it still achieves reasonable performance when the privacy budget is relatively large. Additionally, DPSGD-LR does not require a sampling process during training, making it more efficient and easier to implement with existing tools.
\end{itemize}

\subsection{Ablation Study}
\label{sec:exp_albation}

In this section, we conduct a series of experiments to investigate the key factors affecting the performance of DPSGD-LR and PAC-LR.

\mypara{Impact of Hyperparameter Tuning}
We find that hyperparameter tuning is crucial for achieving reasonable performance for both DPSGD-LR and PAC-LR. 
Using grid search, we select the optimal hyperparameters with the lowest RMSE, as listed in Table~\ref{tab:hyperparams}. 
We also compare the performance of models with default settings against those with tuned hyperparameters, as shown in Table~\ref{tab:ablation_tune}.
For the default, we chose the learning rate as 0.01, batch size as 16, norm clip as 1, epochs as 10, and regularization penalty $\lambda$ as zero across three datasets. 
The results indicate that hyperparameters have a significant impact on performance: both DPSGD-LR and PAC-LR show substantial improvements across all three datasets after tuning. 
This effect is particularly noticeable for DPSGD-LR, which has more hyperparameters to tune. In contrast, PAC-LR has only one hyperparameter (\ie regularization penalty $\lambda$), making it more robust to hyperparameter settings and demonstrating more stable performance.

\mypara{Impact of Data Normalization}
We also investigate the impact of data normalization for privacy-preserving linear regression. 
Specifically, we compare the performance of DPSGD-LR and PAC-LR trained with feature normalization against models trained on raw data, with results recorded in Table~\ref{tab:ablation_norm}.
The results show that without normalization, the performance of both algorithms drops dramatically, and in some cases (\ie the Automobiles dataset), the models fail to converge. 
We attribute this to the DPSGD mechanism: it clips gradients uniformly across all features for privacy accounting. When feature ranges vary significantly, this uniform clipping disrupts gradient updates, making convergence difficult.
Similarly, without data normalization, the output of linear regression becomes more diverse, requiring larger noise to satisfy PAC privacy guarantees, further degrading performance.

\mypara{Impact of Regularization and Projection for PAC-LR}
We evaluate the impact of three commonly used regularization techniques for linear regression: no regularization (Ordinary Least Squares, OLS), $l_2$ regularization (Ridge), and $l_1$ regularization (Lasso). Additionally, we compare the use of a unitary projection matrix with a projection matrix learned from Singular Value Decomposition (SVD). 
The results are shown in Table~\ref{tab:ablation_regularzation}.

The results indicate that regularization improves performance by providing more robust linear regression estimates, leading to improvements across all three datasets. Furthermore, using an SVD-learned projection matrix helps produce smaller anisotropic noise, resulting in better overall performance.

\begin{table}[t]
    \centering
    \caption{Impact of hyperparameter tuning for DPSGD-LR and PAC-LR. We fix the posterior successful rate as 0.75 and use RMSE as the evaluation metric.}
    {\small 

\resizebox{0.44\textwidth}{!}{
\begin{tabular}{cccccc}
\toprule[0.8pt]
& & \textbf{Lenses} & \textbf{Concrete}  & \textbf{Automobiles} \\
\midrule

\multirow{2}{*}{\begin{tabular}[c]{@{}c@{}} DPSGD-LR \end{tabular}} 
& Default & $1.384$ & $1.016$ & $0.584$ 
 \\ 
\cmidrule(lr){2-5} 
& Tuned & $\mathbf{0.971}$ & $\mathbf{0.389}$ & $\mathbf{0.504}$ \\
\midrule
\multirow{2}{*}{\begin{tabular}[c]{@{}c@{}} PAC-LR \end{tabular}} 
& Default & $0.653$ & $0.323$ & $0.398$
 \\ 
\cmidrule(lr){2-5} 
& Tuned & $\mathbf{0.642}$ & $\mathbf{0.321}$ & $\mathbf{0.392}$ \\

\bottomrule
\end{tabular}}

}
    \label{tab:ablation_tune}
\end{table}

\begin{table}[t]
    \centering
    \caption{Impact of normalization for DPSGD-LR and PAC-LR. We fix the posterior successful rate as 0.75 and use RMSE as the evaluation metric. ``-'' means it fails to achieve reasonable performance.}
    {\small \resizebox{0.48\textwidth}{!}{
\begin{tabular}{cccccc}
\toprule[0.8pt]
& & \textbf{Lenses} & \textbf{Concrete}  & \textbf{Automobiles} \\
\midrule
\multirow{2}{*}{\begin{tabular}[c]{@{}c@{}} DPSGD-LR \end{tabular}} 
& w/o normalization & $0.805$ & $10.445$ & - 
 \\ 
\cmidrule(lr){2-5} 
& w/ normalization & $\mathbf{0.681}$ & $\mathbf{0.335}$ & $\mathbf{0.441}$ \\
\midrule
\multirow{2}{*}{\begin{tabular}[c]{@{}c@{}} PAC-LR \end{tabular}} 
& w/o normalization & $0.528$ & $3.375$ & -
 \\ 
\cmidrule(lr){2-5} 
& w/ normalization & $\mathbf{0.634}$ & $\mathbf{0.321}$ & $\mathbf{0.386}$ \\

\bottomrule
\end{tabular}}

}
    \label{tab:ablation_norm}
\end{table}

\begin{table}[t]
    \centering
    \caption{Impact of regularization and projection for PAC-LR. We fix the posterior successful rate as 0.75 and use RMSE as the evaluation metric.}
    {\small \resizebox{0.46\textwidth}{!}{
\begin{tabular}{cccccc}
\toprule[0.8pt]
& & \textbf{Lenses} & \textbf{Concrete}  & \textbf{Automobiles} \\
\midrule
\multirow{2}{*}{\begin{tabular}[c]{@{}c@{}} PAC-OLS \end{tabular}} 
& unitary projection & $0.867$ & $2.735$ & $0.519$ 
 \\ 
\cmidrule(lr){2-5} 
& SVD projection & $\mathbf{0.652}$ & $\mathbf{0.322}$ & $\mathbf{0.398}$ \\
\midrule
\multirow{2}{*}{\begin{tabular}[c]{@{}c@{}} PAC-Ridge \end{tabular}} 
& unitary projection & $0.703$ & $0.321$ & $0.402$
 \\ 
\cmidrule(lr){2-5} 
& SVD projection & $\mathbf{0.639}$ & $\mathbf{0.321}$ & $\mathbf{0.386}$ \\
\midrule
\multirow{2}{*}{\begin{tabular}[c]{@{}c@{}} PAC-Lasso \end{tabular}} 
& unitary projection & $0.761$ & $0.331$ & $0.438$
 \\ 
\cmidrule(lr){2-5} 
& SVD projection & $\mathbf{0.642}$ & $\mathbf{0.322}$ & $\mathbf{0.391}$  \\ 
\bottomrule
\end{tabular}}
}
    \label{tab:ablation_regularzation}
\end{table}

%


\section{Conclusion}

In this paper, we study the problem of training linear regression models with provable privacy guarantees. 
Leveraging two well-established privacy notions (\ie differential privacy and PAC privacy), we adopt two privacy-preserving linear regression models: DPSGD-LR and PAC-LR.
Experiments on three real-world datasets demonstrate the superiority of PAC-LR, particularly under strict privacy guarantees. 
Additionally, we observe that data normalization and regularization significantly impact the performance of both DPSGD-LR and PAC-LR, aspects that have been largely overlooked in previous studies.
Looking forward, we plan to compare PAC-LR with more state-of-the-art DP linear regression methods. 
It would also be interesting to improve the sampling efficiency of PAC privacy for training linear regression models.
Finally, we intend to systematically investigate the role of regularization in learning smaller anisotropic noise in PAC privacy.

\section*{Acknowledgements}
We would like to thank Dr. Ninghui Li of Purdue University for the introduction to and guidance throughout this project, Yuntao Du for his invaluable support during the writing and experimentation process, and Hanshen Xiao for his technical insight and expertise. This research was supported in part by Lilly Endowment, Inc., through its support for the Indiana University Pervasive Technology Institute.

\bibliographystyle{unsrt}
\bibliography{ref}
\end{document}